\documentclass{article}

\usepackage[preprint,nonatbib]{nips_2018}

\usepackage[utf8]{inputenc} % allow utf-8 input
\usepackage[T1]{fontenc}    % use 8-bit T1 fonts
\usepackage{hyperref}       % hyperlinks
\usepackage{url}            % simple URL typesetting
\usepackage{booktabs}       % professional-quality tables
\usepackage{amsmath,amssymb,amsbsy,amsfonts}
\usepackage{nicefrac}       % compact symbols for 1/2, etc.
\usepackage{microtype}      % microtypography
\usepackage{fancyvrb}
\usepackage{yfonts}
\usepackage{color}
\usepackage{wrapfig}
\usepackage{tikz} % To generate the plot from csv
\usepackage{pgfplots}
\usepackage{booktabs} % for professional tables
\usepackage{multirow}
\usepackage[normalem]{ulem}
\useunder{\uline}{\ul}{}
\usepackage[shortlabels]{enumitem}
\usepackage{subfigure}
\usepackage{graphicx,psfrag,epsf}
\usepackage[ruled]{algorithm2e}

\SetCommentSty{commfont}
\SetKwComment{Comment}{~\#~}{}

\newcommand{\R}{\mathbb{R}}

\newcommand{\vect}[1]{\mathbf{#1}}
\newcommand*{\QEDB}{\hfill\ensuremath{\square}}
\DeclareMathOperator*{\argmin}{argmin}
\DeclareMathOperator*{\argmax}{argmax}

\DeclareMathOperator*{\ifn}{if}
\DeclareMathOperator*{\otherwise}{otherwise}

\newtheorem{property}{Property}[section]

\title{Deep $k$-Means: Jointly clustering with $k$-Means and learning representations}

\author{
  Maziar Moradi Fard \\
  Univ. Grenoble Alpes, CNRS, Grenoble INP -- LIG \\
  \texttt{maziar.moradi-fard@univ-grenoble-alpes.fr} \\
  \And
  Thibaut Thonet \\
  Univ. Grenoble Alpes, CNRS, Grenoble INP -- LIG \\
  \texttt{thibaut.thonet@univ-grenoble-alpes.fr} \\
  \And
  Eric Gaussier \\
  Univ. Grenoble Alpes, CNRS, Grenoble INP -- LIG, Skopai \\
  \texttt{eric.gaussier@univ-grenoble-alpes.fr} \\
}

\begin{document}

\maketitle

\begin{abstract}
We study in this paper the problem of jointly clustering and learning representations. As several previous studies have shown, learning representations that are both faithful to the data to be clustered and adapted to the clustering algorithm can lead to better clustering performance, all the more so that the two tasks are performed jointly. We propose here such an approach for $k$-Means clustering based on a continuous reparametrization of the objective function that leads to a truly joint solution. The behavior of our approach is illustrated on various datasets showing its efficacy in learning representations for objects while clustering them.
\end{abstract}

\section{Introduction}
\label{sec:intro}

Clustering is a long-standing problem in the machine learning and data mining fields, and thus accordingly fostered abundant research. Traditional clustering methods, \textit{e.g.}, $k$-Means~\cite{MacQueen1967} and Gaussian Mixture Models (GMMs)~\cite{Bishop2006}, fully rely on the original data representations and may then be ineffective when the data points (\textit{e.g.}, images and text documents) live in a high-dimensional space~-- a problem commonly known as the curse of dimensionality. Significant progress has been made in the last decade or so to learn better, low-dimensional data representations~\cite{Hinton2006}. The most successful techniques to achieve such high-quality representations rely on deep neural networks (DNNs), which apply successive non-linear transformations to the data in order to obtain increasingly high-level features. Auto-encoders (AEs) are a special instance of DNNs which are trained to embed the data into a (usually dense and low-dimensional) vector at the bottleneck of the network, and then attempt to reconstruct the input based on this vector. The appeal of AEs lies in the fact that they are able to learn representations in a fully unsupervised way. 
%Their broad potential and applicability inspired many variants or alternatives: \textit{e.g.}, deep belief networks~\cite{Hinton2006b}, denoising AEs~\cite{Vincent2010}, contractive AEs~\cite{Rifai2011}, and variational AEs~\cite{Kingma2014}.
The representation learning breakthrough enabled by DNNs spurred the recent development of numerous deep clustering approaches which aim at jointly learning the data points' representations as well as their cluster assignments. 

%For that purpose, they usually rely on a minimization problem based on a trade-off between the reconstruction error provided by an auto-encoder and a clustering loss associated with the clustering algorithm used, as:
%%
%\begin{equation}\label{eq:gen-min}
%\underset{\theta}{\min} \, \underset{x \in \mathcal{X}}{\sum} \lambda \, L_{rec}(x) + L_{clust}(x)
%\end{equation}
%%
%where $x$ denotes an object from the set of objects $\mathcal{X}$ to be clustered, $\lambda$ a positive scalar controlling the trade-off between the two terms~-- the AE's reconstruction error ($L_{rec}$) and the clustering loss ($L_{clust}$)~-- and $\theta$ the set of parameters of the model, which $L_{rec}$ and $L_{clust}$ depend on.

In this study, we specifically focus on the $k$-Means-related deep clustering problem. Contrary to previous approaches that alternate between continuous gradient updates and discrete cluster assignment steps~\cite{Yang2017}, we show here that one can solely rely on gradient updates to learn, truly jointly, representations and clustering parameters. This ultimately leads to a better deep $k$-Means method which is also more scalable as it can fully benefit from the efficiency of stochastic gradient descent (SGD). In addition, we perform a careful comparison of different methods by \textit{(a)} relying on the same auto-encoders, as the choice of auto-encoders impacts the results obtained, \textit{(b)} tuning the hyperparameters of each method on a small validation set, instead of setting them without clear criteria, and \textit{(c)} enforcing, whenever possible, that the same initialization and sequence of SGD minibatches are used by the different methods. The last point is crucial to compare different methods as these two factors play an important role and the variance of each method is usually not negligible.

\section{Related work}
\label{sec:relwork}

In the wake of the groundbreaking results obtained by DNNs in computer vision, several deep clustering algorithms were specifically designed for image clustering \cite{Yang2016,Chang2017,Dizaji2017,Hu2017,Hsu2018}. These works have in common the exploitation of Convolutional Neural Networks (CNNs), which extensively contributed to last decade's significant advances in computer vision.
Inspired by agglomerative clustering, \cite{Yang2016} proposed a recurrent process which successively merges clusters and learn image representations based on CNNs. In~\cite{Chang2017}, the clustering problem is formulated as binary pairwise-classification so as to identify the pairs of images which should belong to the same cluster.
Due to the unsupervised nature of clustering, the CNN-based classifier in this approach is only trained on noisily labeled examples obtained by selecting increasingly difficult samples in a curriculum learning fashion.
\cite{Dizaji2017} jointly trained a CNN auto-encoder and a multinomial logistic regression model applied to the AE's latent space. Similarly, \cite{Hsu2018} alternate between representation learning and clustering where mini-batch $k$-Means is utilized as the clustering component. Differently from these works, \cite{Hu2017} proposed an information-theoretic framework based on data augmentation to learn discrete representations, which may be applied to clustering or hash learning.
Although these different algorithms obtained state-of-the-art results on image clustering~\cite{Aljalbout2018}, their ability to generalize to other types of data (e.g., text documents) is not guaranteed due to their reliance on essentially image-specific techniques~-- Convolutional Neural Network architectures and data augmentation. 
%\cite{Guo2017,Li2017}

Nonetheless, many general-purpose~-- non-image-specific~-- approaches to deep clustering have also been recently designed~\cite{Huang2014,Peng2016,Xie2016,Dilokthanakul2017,Guo2017b,Hu2017,Ji2017,Jiang2017,Peng2017,Yang2017}. Generative models were proposed in \cite{Dilokthanakul2017,Jiang2017} which combine variational AEs and GMMs to perform clustering. Alternatively, \cite{Peng2016,Peng2017,Ji2017} framed deep clustering as a subspace clustering problem in which the mapping from the original data space to a low-dimensional subspace is learned by a DNN. \cite{Xie2016} defined the Deep Embedded Clustering (DEC) method which simultaneously updates the data points' representations, initialized from a pre-trained AE, and cluster centers. DEC uses soft assignments which are optimized to match stricter assignments through a Kullback-Leibler divergence loss. IDEC was subsequently proposed in \cite{Guo2017b} as an improvement to DEC by integrating the AE's reconstruction error in the objective function.

Few approaches were directly influenced by $k$-Means clustering \cite{Huang2014,Yang2017}. The Deep Embedding Network (DEN) model \cite{Huang2014} first learns representations from an AE while enforcing locality-preserving constraints and group sparsity; clusters are then obtained by simply applying $k$-Means to these representations. Yet, as representation learning is decoupled from clustering, the performance is not as good as the one obtained by methods that rely on a joint approach. Besides \cite{Hsu2018}, mentioned before in the context of images, the only study, to our knowledge, that directly addresses the problem of jointly learning representations and clustering with $k$-Means (and not an approximation of it) is the Deep Clustering Network (DCN) approach \cite{Yang2017}. However, as in \cite{Hsu2018}, DCN alternatively learns (rather than jointly learns) the object representations, the cluster centroids and the cluster assignments, the latter being based on discrete optimization steps which cannot benefit from the efficiency of stochastic gradient descent. The approach proposed here, entitled \textit{Deep $k$-Means} (DKM), addresses this problem.

\section{Deep k-Means}
\label{sec:deepKmeans}

In the remainder, $x$ denotes an object from a set $\mathcal{X}$ of objects to be clustered. $\R^p$ represents the space in which learned data representations are to be embedded. $K$ is the number of clusters to be obtained, $\vect{r}_k \in \R^p$ the representative of cluster $k, \, 1 \le k \le K$, and $\mathcal{R}= \{\mathbf{r}_1, \ldots, \mathbf{r}_K\}$ the set of representatives. Functions $f$ and $g$ define some distance in $\R^p$ which are assumed to be fully differentiable wrt their variables. For any vector $\vect{y} \in \R^p$, $c_f(\vect{y};\mathcal{R})$ gives the closest representative of $\vect{y}$ according to $f$.

The deep $k$-Means problem takes the following form:
\begin{align}
\label{eq:gen-kmeans}
\begin{aligned}
& \underset{\mathcal{R},\theta}{\min} \, \underset{x \in \mathcal{X}}{\sum} g(x,A(x;\theta)) + \lambda f(\vect{h}_{\theta}(x),c_f(\vect{h}_{\theta}(x);\mathcal{R})), & \\
& \mbox{with:} \,\, c_f(\vect{h}_{\theta}(x);\mathcal{R}) = \underset{\vect{r} \in \mathcal{R}}{\argmin} \, f(\vect{h}_{\theta}(x),\vect{r}) & 
\end{aligned}
\end{align}
$g$ measures the error between an object $x$ and its reconstruction $A(x;\theta)$ provided by an auto-encoder, $\theta$ representing the set of the auto-encoder's parameters. A regularization term on $\theta$ can be included in the definition of $g$. However, as most auto-encoders do not use regularization, we dispense with such a term here. $\vect{h}_{\theta}(x)$ denotes the representation of $x$ in $\R^p$ output by the AE's encoder part and $f(\vect{h}_{\theta}(x),c_f(\vect{h}_{\theta}(x);\mathcal{R}))$ is the clustering loss corresponding to the $k$-Means objective function in the embedding space. Finally, $\lambda$ in Problem (\ref{eq:gen-kmeans}) regulates the trade-off between seeking good representations for $x$~-- \textit{i.e.}, representations that are faithful to the original examples~-- and representations that are useful for clustering purposes. Similar optimization problems can be formulated when $f$ and $g$ are similarity functions or a mix of similarity and distance functions. The approach proposed here directly applies to such cases.
%
%The sign variables $\epsilon_0$ and $\epsilon_1$ are set as follows:
%
%\begin{equation}\label{eq:esp0}
%\epsilon_0 \, (\mbox{resp. } \epsilon_1) =
%\begin{cases}
% +1 \, \mbox{if $g$ (resp. $f$) is a distance,} \nonumber \\
% -1 \, \mbox{if $g$ (resp. $f$) is a similarity function.} %\nonumber
%\end{cases}
%\end{equation}
%
%By \textit{same nature}, we mean here that they are both similarity or both dissimilarity functions. For $\epsilon_1$:
%%
%\begin{equation}\label{eq:esp0}
%\epsilon_1 =
%\begin{cases}
% +1 \, \mbox{if $f$ is a dissimilarity function,} \nonumber \\
% -1 \, \mbox{otherwise.} \nonumber
%\end{cases}
%\end{equation}
%%
%Standard $k$-Means is obtained by setting $f$ to the Euclidean distance. In the remainder, we focus on functions $g$ that are differentiable wrt $\theta$ and functions $f$ that are differentiable wrt to both $\theta$ and $\mathcal{R}$, where differentiability wrt $\mathcal{R}$ means differentiability wrt to all dimensions of $\mathbf{r}_k, \, 1 \le k \le K$. 

\begin{figure}[t]
\centering
\includegraphics[width=10cm]{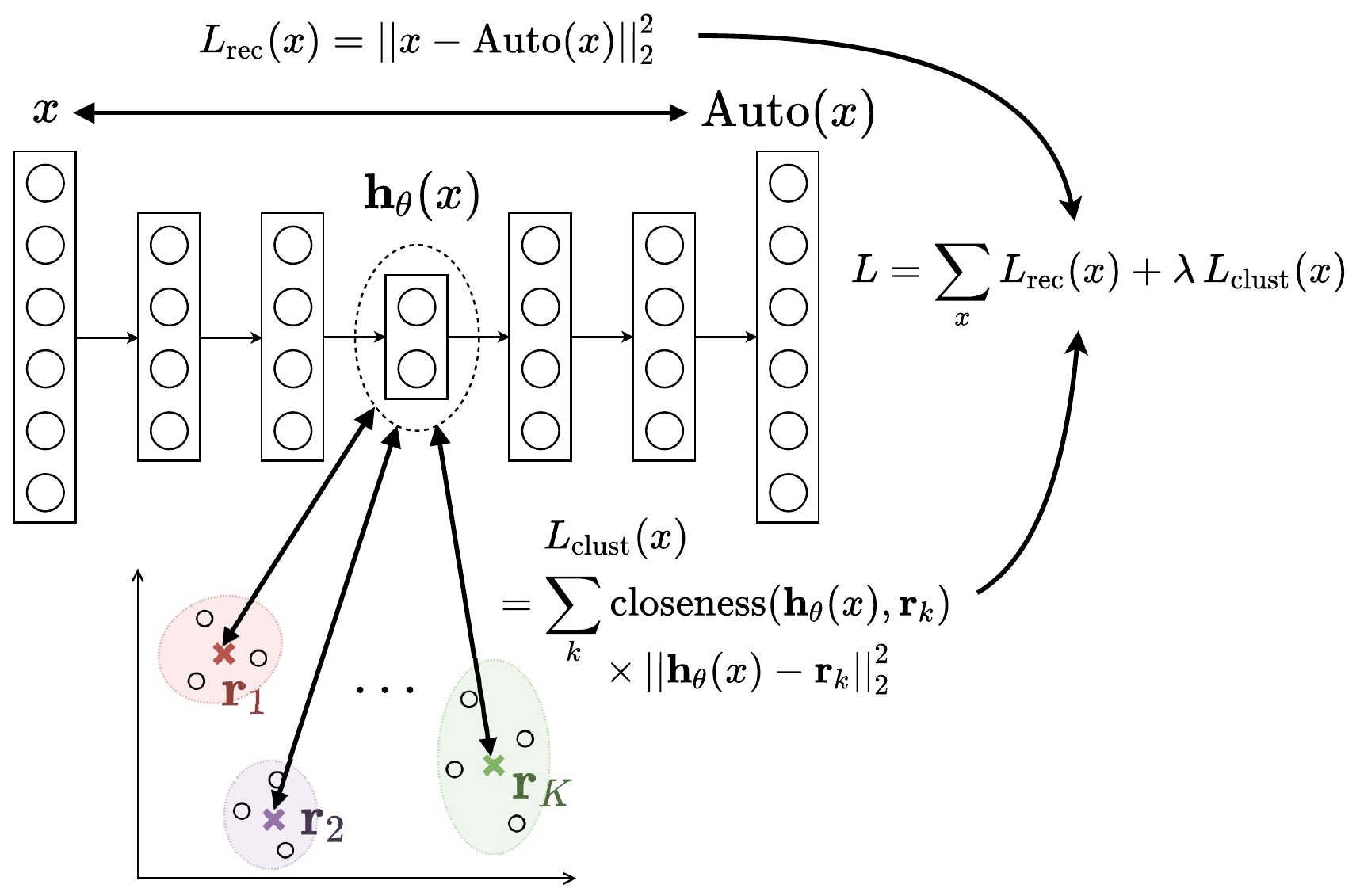}
\caption{Overview of the proposed Deep k-Means approach instantiated with losses based on the Euclidean distance.}
\label{fig:illus}
\end{figure}

Figure~\ref{fig:illus} illustrates the overall framework retained in this study with $f$ and $g$ both based on the Euclidean distance. The closeness term in the clustering loss will be further clarified below. 

\subsection{Continuous generalization of Deep $k$-Means} 

We now introduce a parameterized version of the above problem that constitutes a \textit{continuous generalization}, whereby we mean here that all functions considered are continuous wrt the introduced parameter.\footnote{Note that, independently from this work, a similar relaxation has been previously proposed in \cite{Agustsson2017}~-- wherein soft-to-hard quantization is performed on an embedding space learned by an AE for compression. However, given the different nature of the goal here~-- clustering~-- our proposed learning framework substantially differs from theirs.} To do so, we first note that the clustering objective function can be rewritten as $f(\vect{h}_{\theta}(x),c_f(\vect{h}_{\theta}(x);\mathcal{R})) = \sum_{k=1}^K f_k(\vect{h}_{\theta}(x);\mathcal{R})$ 
%%
%\[
%f(\vect{h}_{\theta}(x),c_f(\vect{h}_{\theta}(x);\mathcal{R})) = \sum_{k=1}^K f_k(\vect{h}_{\theta}(x);\mathcal{R})
%\]
%%
with:
\begin{equation}\label{eq:fk}
f_k(\vect{h}_{\theta}(x);\mathcal{R}) = 
\begin{cases}
f(\vect{h}_{\theta}(x),\mathbf{r}_k) & \ifn \, \mathbf{r}_k = c_f(\vect{h}_{\theta}(x);\mathcal{R}) \nonumber \\
0 & \otherwise \nonumber
\end{cases}
\end{equation}
Let us now assume that we know some function $G_{k,f}(\vect{h}_{\theta}(x),\alpha;\mathcal{R})$ such that:
\begin{enumerate}[(i),leftmargin=0.7cm]
\item $G_{k,f}$ is differentiable wrt to $\theta, \mathcal{R}$ and continuous wrt $\alpha$ (differentiability wrt $\mathcal{R}$ means differentiability wrt to all dimensions of $\mathbf{r}_k, \, 1 \le k \le K$);
\item $\exists \alpha_0 \in \mathbb{R} \cup \{-\infty,+\infty\}$ such that:
\begin{equation}\label{eq:Galpha}
\underset{\alpha \rightarrow \alpha_0}{\lim} G_{k,f}(\vect{h}_{\theta}(x),\alpha;\mathcal{R}) = 
\begin{cases}
1 & \ifn \, \mathbf{r}_k = c_f(\vect{h}_{\theta}(x);\mathcal{R}) \nonumber \\
0 & \otherwise \nonumber
\end{cases}
\end{equation} 
\end{enumerate}

Then, one has, $\forall x \in \mathcal{X}$: $\underset{\alpha \rightarrow \alpha_0}{\lim} f(\vect{h}_{\theta}(x),\mathbf{r}_k) \, G_{k,f}(\vect{h}_{\theta}(x),\alpha;\mathcal{R}) = f_k(\vect{h}_{\theta}(x);\mathcal{R})$, showing that the problem in (\ref{eq:gen-kmeans}) is equivalent to:
\begin{equation}\label{eq:alphagenkmeans}
\underset{\mathcal{R},\theta}{\min} \,  \underset{\alpha \rightarrow \alpha_0}{\lim} \overbrace{\sum_{x \in \mathcal{X}} g(x,A(x;\theta)) \, + \lambda \sum_{k=1}^K f(\vect{h}_{\theta}(x),\mathbf{r}_k) \, G_{k,f}(\vect{h}_{\theta}(x),\alpha;\mathcal{R})}^{\mathcal{F}(\mathcal{X},\alpha;\theta,\mathcal{R})}
\end{equation}
%
%Then, one has:
%%
%\begin{lemma}
%\label{lem:falpha}
%$\forall x \in \mathcal{X}$:
%%
%\[
%\underset{\alpha \rightarrow \alpha_0}{\lim} f(\vect{h}_{\theta}(x),\mathbf{r}_k) \, G_{k,f}(\vect{h}_{\theta}(x),\alpha;\mathcal{R}) = f_k(\vect{h}_{\theta}(x);\mathcal{R})
%\]
%%
%\end{lemma}
%%
%The proof directly derives from the definitions of $f_k$ and $G_{k,f}$.
%
%From Lemma~\ref{lem:falpha}, one can see that:
%%
%\begin{property}
%\label{prop:equiv-genkmeans}
%The problem given in (\ref{eq:gen-kmeans}) is equivalent to:
%%
%\begin{align}\label{eq:alphagenkmeans}
%%\underset{\mathcal{R},\theta}{\min} \,  \underset{\alpha \rightarrow \alpha_0}{\lim} \overbrace{\sum_{x \in \mathcal{X}} \epsilon_0 g(x,A(x;\theta)) \, + \lambda \epsilon_1 \sum_{k=1}^K f(\vect{h}_{\theta}(x),\mathbf{r}_k) \, G_{k,f}(\vect{h}_{\theta}(x),\alpha;\mathcal{R})}^{\mathcal{F}(\mathcal{X},\alpha;\theta,\mathcal{R})}
%& \underset{\mathcal{R},\theta}{\min} \,  \underset{\alpha \rightarrow \alpha_0}{\lim} \mathcal{F}(\mathcal{X},\alpha;\theta,\mathcal{R}), & \nonumber \\
%& \mbox{with:} \, \mathcal{F}(\mathcal{X},\alpha;\theta,\mathcal{R}) = \left( \sum_{x \in \mathcal{X}} \epsilon_0 g(x,A(x;\theta)) \, + \right. & \nonumber \\
%& \hspace{1cm} \left. \lambda \epsilon_1 \sum_{k=1}^K f(\vect{h}_{\theta}(x),\mathbf{r}_k) \, G_{k,f}(\vect{h}_{\theta}(x),\alpha;\mathcal{R}) \right) & 
%\end{align}
%%
%\end{property}
%
All functions in the above formulation are fully differentiable wrt both $\theta$ and $\mathcal{R}$. One can thus estimate $\theta$ and $\mathcal{R}$ through a simple, joint optimization based on stochastic gradient descent (SGD) for a given $\alpha$:
\begin{equation}\label{eq:update}
(\theta, \, \mathcal{R}) \leftarrow (\theta, \, \mathcal{R}) - \eta \, \frac{1}{|\tilde{\mathcal{X}}|} \, \nabla_{(\theta, \, \mathcal{R})} \mathcal{F}(\tilde{\mathcal{X}},\alpha;\theta,\mathcal{R})
\end{equation}
with $\eta$ the learning rate and $\tilde{\mathcal{X}}$ a random mini-batch of $\mathcal{X}$.
% COMMENT: Added the stochastic version here as we only had gradient descent before (F is based on full batch).

\subsection{Choice of ${G_{k,f}}$}

Several choices are possible for $G_{k,f}$. A simple choice, used throughout this study, is based on a parameterized softmax function. The fact that the softmax function can be used as a differentiable surrogate to $\argmax$ or $\argmin$ is well known and has been applied in different contexts, as in the recently proposed Gumbel-softmax distribution employed to approximate categorical samples \cite{Jang2017,Maddison2017}. The parameterized softmax function which we adopted takes the following form:  
\begin{equation}\label{eq:G1}
G_{k,f}(\vect{h}_{\theta}(x),\alpha;\mathcal{R}) = \frac{e^{-\alpha f(\vect{h}_{\theta}(x),\mathbf{r}_k)}}{\sum_{k'=1}^K e^{-\alpha f(\vect{h}_{\theta}(x),\mathbf{r}_{k'})}}
\end{equation}
%\begin{equation}\label{eq:G1}
%G_{k,f}(\vect{h}_{\theta}(x),\alpha;\mathcal{R}) = \frac{e^{-\epsilon_1 \alpha f(\vect{h}_{\theta}(x),\mathbf{r}_k)}}{\sum_{k'=1}^K e^{-\epsilon_1 \alpha f(\vect{h}_{\theta}(x),\mathbf{r}_{k'})}}
%\end{equation}
%
with $\alpha \in [0,+\infty)$. The function $G_{k,f}$ defined by Eq.~\ref{eq:G1} is differentiable wrt $\theta, \mathcal{R}$ and $\alpha$ (condition \textit{(i)}) as it is a composition of functions differentiable wrt these variables. Furthermore, one has:
\begin{property}(condition \textit{(ii)})\label{prop-f-alpha} If $c_f(\vect{h}_{\theta}(x);\mathcal{R})$ is unique for all $x \in \mathcal{X}$, then:
\begin{equation}\label{eq:propG1}
\underset{\alpha \rightarrow +\infty}{\lim} \, G_{k,f}(\vect{h}_{\theta}(x),\alpha;\mathcal{R}) = 
\begin{cases}
1 \, \ifn \, \mathbf{r}_k = c_f(\vect{h}_{\theta}(x);\mathcal{R}) \nonumber \\
0 \, \otherwise \nonumber
\end{cases}
\end{equation}
\end{property}
The proof, which is straightforward, is detailed in the Supplementary Material.
%\textbf{Proof:} Let $\mathbf{r}_j = c_f(\vect{h}_{\theta}(x);\mathcal{R})$ and let us assume that $f$ is a distance. One has:
%%
%\begin{align}\label{eq:proof1}
%\sum_{k'=1}^K & e^{- \alpha f(\vect{h}_{\theta}(x),\mathbf{r}_{k'})} = e^{- \alpha f(\vect{h}_{\theta}(x),\mathbf{r}_j)} \nonumber \\
%& \times \left( 1+ \sum_{k' \ne j} e^{-\alpha (f(\vect{h}_{\theta}(x),\mathbf{r}_{k'}) - f(\vect{h}_{\theta}(x),\mathbf{r}_j))} \right) \nonumber
%\end{align}
%%\begin{align}\label{eq:proof1}
%%\sum_{k'=1}^K e^{- \alpha f(\vect{h}_{\theta}(x),\mathbf{r}_{k'})} = e^{- \alpha f(\vect{h}_{\theta}(x),\mathbf{r}_j)} \nonumber\times \left( 1+ \sum_{k' \ne j} e^{-\alpha (f(\vect{h}_{\theta}(x),\mathbf{r}_{k'}) - f(\vect{h}_{\theta}(x),\mathbf{r}_j))} \right) \nonumber
%%\end{align}
%%
%As $f(\vect{h}_{\theta}(x),\mathbf{r}_j) < f(\vect{h}_{\theta}(x),\mathbf{r}_{k'}), \, \forall k' \ne j$, one has:
%%
%\[
%\underset{\alpha \rightarrow +\infty}{\lim} e^{-\alpha (f(\vect{h}_{\theta}(x),\mathbf{r}_{k'}) - f(\vect{h}_{\theta}(x),\mathbf{r}_j))} =0
%\]
%%
%Thus $\underset{\alpha \rightarrow +\infty}{\lim} G_{k,f}(\vect{h}_{\theta}(x),\alpha;\mathcal{R}) = 0$ if $k \ne j$ and $1$ if $k=j$. $\QEDB$

The assumption that $c_f(\vect{h}_{\theta}(x);\mathcal{R})$ is unique for all objects is necessary for $G_{k,f}$ to take on binary values in the limit; it is not necessary to hold for small values of $\alpha$.
%In fact, when $\alpha$ is set to $0$, the deep $k$-Means optimization problem has a unique solution obtained by setting all representatives to the (single) vector\footnote{This vector corresponds to the centroid of all embedded representations when $f$ is the Euclidean distance.} that optimizes $\sum_{x \in \mathcal{X}} f(\vect{h}_{\theta}(x),\mathbf{r})$. 
In the unlikely event that the above assumption does not hold for some $x$ and large $\alpha$, one can slightly perturbate the representatives equidistant to $x$ prior to updating them. We have never encountered this situation in practice. 

Finally, Eq.~\ref{eq:G1} defines a valid (according to conditions $(i)$ and $(ii)$) function $G_{k,f}$ that can be used to solve the deep $k$-Means problem (\ref{eq:alphagenkmeans}). We adopt this function in the remainder of this study. 
%Note that the membership function of the fuzzy $C$-Means algorithm \cite{Bezdek1984} is a also valid candidate according to conditions $(i)$ and $(ii)$. However, in addition to being slightly more complex than the parametrized softmax, this formulation presents the disadvantage that it may be undefined when a representative coincides with an object.

%Prior to studying the effect of $\alpha$ in Eq.~\ref{eq:G1}, we want to mention another possible choice for $G_{k,f}$. As $G_{k,f}(\vect{h}_{\theta}(x),\alpha;\mathcal{R})$ plays the role of a closeness function for object $x$ wrt representative $\vect{r}_k$, membership functions used in fuzzy clustering are potential candidates for $G_{k,f}$. In particular, the membership function of the fuzzy $C$-Means algorithm \cite{Bezdek1984} is a valid candidate according to conditions $(i)$ and $(ii)$. It takes the form, for distance functions:
%%
%\[
%G_{k,f}(\vect{h}_{\theta}(x),\alpha;\mathcal{R}) = \left( \sum_{k'=1}^K \left(\frac{f(\vect{h}_{\theta}(x),\mathbf{r}_k)}{f(\vect{h}_{\theta}(x),\mathbf{r}_{k'})}\right)^{\frac{2}{\alpha-1}} \right)^{-1}
%\]
%%
%with $\alpha$ defined on $(1;+\infty)$ and $\alpha_0$ (condition $(ii)$) equal to 1. However, in addition to being slightly more complex than the parametrized softmax, this formulation presents the disadvantage that it may be undefined when a representative coincides with an object; another assumption (in addition to the uniqueness assumption) is required here to avoid such a case.

\subsection{Choice of $\alpha$}

The parameter $\alpha$ can be defined in different ways. Indeed, $\alpha$ can play the role of an inverse temperature such that, when $\alpha$ is $0$, each data point in the embedding space is equally close, through $G_{k,f}$, to all the representatives (corresponding to a completely soft assignment), whereas when $\alpha$ is $+\infty$, the assignment is hard. In the first case, for the deep $k$-Means optimization problem, all representatives are equal and set to the point $\vect{r} \in \R^p$ that minimizes $\sum_{x \in \mathcal{X}} f(\vect{h}_{\theta}(x),\vect{r})$. In the second case, the solution corresponds to exactly performing $k$-Means in the embedding space, the latter being learned jointly with the clustering process.
% COMMENT: exactly performing deep $k$-Means or only $k$-Means? In case of hard assignments, shouldn't it be $k$-Means? Otherwise I don't fully understand the sentence.
%In such a situation, a deterministic annealing process can be employed to try and keep track of "good" local optima and be less dependent on the initialization of the representatives. This process consists in starting
%
Following a deterministic annealing approach \cite{rose-90}, one can start with a low value of $\alpha$ (close to 0), and gradually increase it till a sufficiently large value is obtained. At first, representatives are randomly initialized. As the problem is smooth when $\alpha$ is close to 0, different initializations are likely to lead to the same local minimum in the first iteration; this local minimum is used for the new values of the representatives for the second iteration, and so on. The continuity of $G_{k,f}$ wrt $\alpha$ implies that, provided the increment in $\alpha$ is not too important, one evolves smoothly from the initial local minimum to the last one.
In the above deterministic annealing scheme, $\alpha$ allows one to initialize cluster representatives. The initialization of the auto-encoder can as well have an important impact on the results obtained and prior studies (\textit{e.g.}, \cite{Huang2014,Xie2016,Guo2017b,Yang2017}) have relied on pretraining for this matter. In such a case, one can choose a high value for $\alpha$ to directly obtain the behavior of the $k$-Means algorithm in the embedding space after pretraining. We evaluate both approaches in our experiments.% (Section~\ref{sec:exps}).

%The development above suggests that one can choose a high value of $\alpha$ to have a behavior that approximates the $k$-Means algorithm in the embedding space. If this approach is feasible, it raises the problem of how to accurately initialize the representatives in the embedding space as such an initialization can have an important impact on the results obtained. Prior studies (\textit{e.g.}, \cite{Huang2014,Xie2016,Guo2017b,Yang2017}) rely on pre-training for this matter, leading to a somewhat cumbersome approach; furthermore, the sensitivity to the original initialization can still be important (see Section~\ref{sec:exps}). 
%%Inspired by deterministic annealing approaches \cite{rose-90}, we propose here to rely on a scheme that starts with low values of $\alpha$ and gradually improves it till convergence, \textit{i.e.}, till $G_{k,f}$ yields binary values (or values deemed sufficiently close to $0$ or $1$).
%We propose here to rely on a deterministic annealing approach \cite{rose-90}.

Algorithm~\ref{algo:train} summarizes the deep $k$-Means algorithm for the deterministic annealing scheme, where $m_{\alpha}$ (respectively $M_{\alpha}$) denote the minimum (respectively maximum) value of $\alpha$, and $T$ is the number of epochs per each value of $\alpha$ for the stochastic gradient updates. 
Even though $M_{\alpha}$ is finite, it can be set sufficiently large to obtain in practice a hard assignment to representatives. Alternatively, when using pretraining, one sets $m_{\alpha} = M_{\alpha}$ (\textit{i.e.}, a constant $\alpha$ is used).

\begin{algorithm}[t!]
\caption{Deep $k$-Means algorithm}
\label{algo:train}
\DontPrintSemicolon
    \KwIn{data $\mathcal{X}$, number of clusters $K$, balancing parameter $\lambda$, scheme for $\alpha$, number of epochs $T$, number of minibatches $N$, learning rate $\eta$}
    %$\mathcal{X}, \lambda, \eta, T, m_{\alpha}, M_{\alpha}$}
    \KwOut{autoencoder parameters $\theta$, cluster representatives $\mathcal{R}$}
    %$\theta$, $\mathcal{R}$}
	%\STATE{\textbf{Random initialization of} $\theta$ \textbf{and} $\vect{r}_k, \, 1 \le k \le K$}
	Initialize $\theta$ and $\vect{r}_k, \, 1 \le k \le K$ (randomly or through pretraining)\;
	\For(\Comment*[f]{inverse temperature}){$\alpha = m_{\alpha}$ {\normalfont \textbf{to}} $M_{\alpha}$} {
		\For(\Comment*[f]{epochs per $\alpha$}){$t=1$ {\normalfont \textbf{to}} $T$} {
			\For(\Comment*[f]{minibatches}){$n=1$ {\normalfont \textbf{to}} $N$} {
				Draw a minibatch $\tilde{\mathcal{X}} \subset \mathcal{X}$\;
				Update $(\theta, \, \mathcal{R})$ using SGD (Eq.~\ref{eq:update})\;
			}
		}
	}
\end{algorithm}

\subsection{Shrinking phenomenon}

The loss functions defined in~\ref{eq:gen-kmeans} and \ref{eq:alphagenkmeans}~-- as well as the loss used in the DCN approach~\cite{Yang2017} and potentially in other approaches~-- might in theory induce a degenerative behavior in the learning procedure. Indeed, the clustering loss could be made arbitrarily small while preserving the reconstruction capacity of the AE by ``shrinking'' the subspace where the object embeddings and the cluster representatives live~-- thus reducing the distance between embeddings and representatives.  %We however noted in practice that this phenomenon was not overwhelming as it did not prevent the model from learning relevant clusters, which we further detail in the Experiments section.
We tested L2 regularization on the auto-encoder parameters to alleviate this potential issue by preventing the weights from arbitrarily shrinking the embedding space (indeed, by symmetry of the encoder and decoder, having small weights in the encoder, leading to shrinking, requires having large weights in the decoder for reconstruction; L2 regularization penalizes such large weights). We have however not observed any difference in our experiments with the case where no regularization is used, showing that the shrinking problem may not be important in practice.  For the sake of simplicity, we dispense with it in the remainder.

%\subsection{Setting $\lambda$}
%\label{sec:lambda}
%
%As mentioned before, $\lambda$ controls the trade-off between learning representations faithful to the data and useful for clustering purposes. It has been proposed in previous studies, as \cite{Yang2017}, to learn it (as well as hyperparameters of the auto-encoder) on a validation set consisting of \textit{labeled} examples. We however want to rely on fully \textit{unsupervised} approaches here and propose to learn $\lambda$ on a validation set consisting of \textit{unlabeled} examples. Indeed, as the two parts, reconstruction and clustering losses, in (\ref{eq:gen-kmeans}) correspond to unrelated objectives, a good trade-off between the two is obtained on that value of $\lambda$ for which the overall objective function is optimal on some validation datasets not used to train the parameters $\mathcal{R}$ and $\theta$. One can thus rely on a line search to find the value of $\lambda$. In practice, we randomly select $10$\% of the \textit{unlabeled} dataset to be clustered as validation set, using the rest as training set. Once the optimal value for $\lambda$ has been found, the whole dataset is used to re-estimate $\mathcal{R}$ and $\theta$.

\section{Experiments}
\label{sec:exps}

In order to evaluate the clustering results of our approach, we conducted experiments on different datasets and compared it against state-of-the-art standard and $k$-Means-related deep clustering models. %We first introduce in Section~\ref{sec:datasets} the datasets we adopted. Then, Section~\ref{sec:train} describes the set of methods which are compared in the experiments.

%\hspace{-0.3cm}
\subsection{Datasets}

The datasets used in the experiments are standard clustering benchmark collections. We considered both image and text datasets to demonstrate the general applicability of our approach. Image datasets consist of \textbf{MNIST} (70,000 images, $28\times28$ pixels, 10 classes) and \textbf{USPS} (9,298 images, $16 \times 16$ pixels, 10 classes) which both contain hand-written digit images. We reshaped the images to one-dimensional vectors and normalized the pixel intensity levels (between 0 and 1 for MNIST, and between -1 and 1 for USPS). The text collections we considered are the 20 Newsgroups dataset (hereafter, \textbf{20NEWS}) and the RCV1-v2 dataset (hereafter, \textbf{RCV1}). For 20NEWS, we used the whole dataset comprising 18,846 documents labeled into 20 different classes. Similarly to~\cite{Xie2016,Guo2017b}, we sampled from the full RCV1-v2 collection a random subset of 10,000 documents, each of which pertains to only one of the four largest classes. Because of the text datasets' sparsity, and as proposed in \cite{Xie2016}, we selected the 2000 words with the highest tf-idf values to represent each document.

%\hspace{-0.3cm}
\subsection{Baselines and deep $k$-Means variants}
\label{sec:baselines}

Clustering models may use different strategies and different clustering losses, leading to different properties. As our goal in this work is to study the $k$-Means clustering algorithm in embedding spaces, we focus on the family of $k$-Means-related models and compare our approach against state-of-the-art models from this family, using both standard and deep clustering models. For the standard clustering methods, we used: the $k$-Means clustering approach \cite{MacQueen1967} with initial cluster center selection \cite{Arthur2007}, denoted \textbf{KM}; an approach denoted as \textbf{AE-KM} in which dimensionality reduction is first performed using an auto-encoder followed by $k$-Means applied to the learned representations.\footnote{We did not consider variational auto-encoders~\cite{Kingma2014} in our baselines as \cite{Jiang2017} previously compared variational AE + GMM and ``standard'' AE + GMM, and found that the latter consistently outperformed the former.} We compared as well against the only previous, ``true'' deep clustering $k$-Means-related method, the Deep Clustering Network (DCN) approach described in~\cite{Yang2017}. DCN is, to the best of our knowledge, the current most competitive clustering algorithm among $k$-Means-related models.

In addition, we consider here the Improved Deep Embedded Clustering (IDEC) model \cite{Guo2017b} as an additional baseline. IDEC is a general-purpose state-of-the-art approach in the deep clustering family. It is an improved version of the DEC model~\cite{Xie2016} and thus constitutes a strong baseline. For both DCN and IDEC, we studied two variants: with pretraining (\textbf{DCN$^\text{p}$} and \textbf{IDEC$^\text{p}$}) and without pretraining (\textbf{DCN$^\text{np}$} and \textbf{IDEC$^\text{np}$}). The pretraining we performed here simply consists in initializing the weights by training the auto-encoder on the data to minimize the reconstruction loss in an end-to-end fashion~-- greedy layer-wise pretraining~\cite{Bengio2006} did not lead to improved clustering in our preliminary experiments.

%\begin{figure}
%\centering
%	\scalebox{0.65}{
%    \begin{tikzpicture}
%      \begin{axis}[
%          width=1.5\linewidth, % Scale the plot to \linewidth
%          grid=major, % Display a grid
%          grid style={dashed,gray!30}, % Set the style
%          xlabel=n, % Set the labels
%          ylabel=$\alpha_n$,
%          x tick label style={anchor=north}, % Display labels sideways
%          y label style={at={(0.1,0.5)}},
%          x label style={at={(0.5,0.02)}},
%        ]
%        \addplot[color=blue]
%        % add a plot from table; you select the columns by using the actual name in the .csv file (on top)
%        table[x expr=\coordindex+1,y=alpha,col sep=comma] {data/dkm_annealed_alpha.csv}; 
%      \end{axis}
%    \end{tikzpicture}
%    }
%    %\vspace{-0.3cm}
%\caption{Annealing scheme for inverse temperature $\alpha$, following the sequence $\alpha_{n+1} = 2^{1/\log(n)^2} \times \alpha_{n}$; $\alpha_1 = 0.1$.}
%\label{fig:alpha_scheme}
%\end{figure}

The proposed Deep $k$-Means (DKM) is, as DCN, a ``true'' $k$-Means approach in the embedding space; it jointly learns AE-based representations and relaxes the $k$-Means problem by introducing a parameterized softmax as a differentiable surrogate to $k$-Means argmin. In the experiments, we considered two variants of this approach. \textbf{DKM$^\text{a}$} implements an annealing strategy for the inverse temperature $\alpha$ and does not rely on pretraining. The scheme we used for the evolution of the inverse temperature $\alpha$ in DKM$^\text{a}$ is given by the following recursive sequence: $\alpha_{n+1} = 2^{1/\log(n)^2} \times \alpha_{n}$ with $m_{\alpha} = \alpha_1 = 0.1$. %The 40 first terms of $(\alpha_{n})$ are plotted in Figure~\ref{fig:alpha_scheme}.
The rationale behind the choice of this scheme is that we want $\alpha$ to spend more iterations on smaller values and less on larger values while preserving a gentle slope. Alternatively, we studied the variant \textbf{DKM$^\text{p}$} which is initialized by pretraining an auto-encoder and then follows Algorithm~\ref{algo:train} with a constant $\alpha$ such that $m_{\alpha} = M_{\alpha} = 1000$. Such a high $\alpha$ is equivalent to having hard cluster assignments while maintaining the differentiability of the optimization problem.

%\hspace{-0.3cm}
\paragraph{Implementation details.} 

For IDEC, we used the Keras code shared by their authors.\footnote{\url{https://github.com/XifengGuo/IDEC-toy}. We used this version instead of \url{https://github.com/XifengGuo/IDEC} as only the former enables auto-encoder pretraining in a non-layer-wise fashion.} Our own code for DKM is based on TensorFlow. To enable full control of the comparison between DCN and DKM~-- DCN being the closest competitor to DKM~-- we also re-implemented DCN in TensorFlow. The code for both DKM and DCN is available online.\footnote{\url{https://github.com/MaziarMF/deep-k-means}}

%\hspace{-0.3cm}
\paragraph{Choice of $f$ and $g$.} 
The functions $f$ and $g$ in Problem~(\ref{eq:gen-kmeans}) define which distance functions is used for the clustering loss and reconstruction error, respectively. In this study, both $f$ and $g$ are simply instantiated with the Euclidean distance on all datasets. For the sake of comprehensiveness, we report in the supplementary material results for the cosine distance on 20NEWS. 
% TODO loss function (f and g are euclidian distance or cosine similarity)

%\begin{figure}[t]
%\centering
%\includegraphics[scale=0.34]{fig/alpha_scheme.pdf}
%\caption{Scheme adopted for the evolution of the inverse temperature $\alpha$, following the sequence $\alpha_{n+1} = 2^{1/\log(n)^2} \, \alpha_{n}$; $\alpha_1 = 0.1$.}
%\label{fig:alpha_scheme}
%\end{figure}

\subsection{Experimental setup}
%\label{sec:exp_setup}

%\hspace{-0.3cm}
\paragraph{Auto-encoder description and training details.} 
The auto-encoder we used in the experiments is the same across all datasets and is borrowed from previous deep clustering studies~\cite{Xie2016,Guo2017b}. Its encoder is a fully-connected multilayer perceptron with dimensions $d$-500-500-2000-$K$, where $d$ is the original data space dimension and $K$ is the number of clusters to obtain. The decoder is a mirrored version of the encoder. All layers except the one preceding the embedding layer and the one preceding the output layer are applied a ReLU activation function \cite{Nair2010} before being fed to the next layer. For the sake of simplicity, we did not rely on any complementary training or regularization strategies such as batch normalization or dropout. The auto-encoder weights are initialized following the Xavier scheme~\cite{Glorot2010}. For all deep clustering approaches, the training is based on the Adam optimizer~\cite{Kingma2015} with standard learning rate $\eta = 0.001$ and momentum rates $\beta_1 = 0.9$ and $\beta_2 = 0.999$. The minibatch size is set to 256 on all datasets following \cite{Guo2017b}. We emphasize that we chose exactly the same training configuration for all models to facilitate a fair comparison.

The number of pretraining epochs is set to 50 for all models relying on pretraining. The number of fine-tuning epochs for DCN$^\text{p}$ and IDEC$^\text{p}$ is fixed to 50 (or equivalently in terms of iterations: 50 times the number of minibatches). %We set the number of training epochs for DCN$^\text{np}$ and IDEC$^\text{np}$ to 50 and 200, respectively~-- in the case of DCN$^\text{np}$, we observed little to no improvement in successive epochs and therefore limited its number of epochs.
We set the number of training epochs for DCN$^\text{np}$ and IDEC$^\text{np}$ to 200. For DKM$^\text{a}$, we used the 40 terms of the sequence $\alpha$ described in Section~\ref{sec:baselines} as the annealing scheme and performed 5 epochs for each $\alpha$ term (\textit{i.e.}, 200 epochs in total). DKM$^\text{p}$ is fine-tuned by performing 100 epochs with constant $\alpha = 1000$. The cluster representatives are initialized randomly from a uniform distribution $U(-1,1)$ for models without pretraining. In case of pretraining, the cluster representatives are initialized by applying $k$-Means to the pretrained embedding space.

%\hspace{-0.3cm}
\paragraph{Hyperparameter selection.} 
The hyperparameters $\lambda$ for DCN and DKM and $\gamma$ for IDEC, that define the trade-off between the reconstruction and the clustering error in the loss function, were determined by performing a line search on the set $\{10^i \, | \, i \in [-4, 3]\}$. To do so, we randomly split each dataset into a validation set (10\% of the data) and a test set (90\%). Each model is trained on the whole data and only the validation set labels are leveraged in the line search to identify the optimal $\lambda$ or $\gamma$ (optimality is measured with respect to the clustering accuracy metric). We provide the validation-optimal $\lambda$ and $\gamma$ obtained for each model and dataset in the supplementary material. The performance reported in the following sections corresponds to the evaluation performed only on the \textit{held-out test set}.

While one might argue that such procedure affects the unsupervised nature of the clustering approaches, we believe that a clear and transparent hyperparameter selection methodology is preferable to a vague or hidden one. Moreover, although we did not explore such possibility in this study, it might be possible to define this trade-off hyperparameter in a data-driven way.

%\hspace{-0.3cm}
\paragraph{Experimental protocol.}

We observed in pilot experiments that the clustering performance of the different models is subject to non-negligible variance from one run to another. This variance is due to the randomness in the initialization  and in the minibatch sampling for the stochastic optimizer. When pretraining is used, the variance of the general pretraining phase and that of the model-specific fine-tuning phase add up, which makes it difficult to draw any confident conclusion about the clustering ability of a model. %Indeed, we noticed that the quality of pretraining often had a major responsibility in the final results.
To alleviate this issue, we compared the different approaches using seeded runs whenever this was possible. This has the advantage of removing the variance of pretraining as seeds guarantee exactly the same results at the end of pretraining (since the same pretraining is performed for the different models). Additionally, it ensures that the same sequence of minibatches will be sampled. In practice, we used seeds for the models implemented in TensorFlow (KM, AE-KM, DCN and DKM). Because of implementation differences, seeds could not give the same pretraining states in the Keras-based IDEC. All in all, we randomly selected 10 seeds and for each model performed one run per seed. Additionally, to account for the remaining variance and to report statistical significance, we performed a Student's $t$-test from the 10 collected samples (i.e., runs).

\begin{table*}[t!]
\centering
\caption{Clustering results of the $k$-Means-related methods. Performance is measured in terms of NMI and ACC (\%); higher is better. Each cell contains the average and standard deviation computed over 10 runs. Bold (resp. underlined) values correspond to results with no significant difference ($p > 0.05$) to the best approach with (resp. without) pretraining for each dataset/metric pair.}
\label{tab:km-results}
\vskip 0.15in
\scalebox{0.87}{
\begin{tabular}{@{}lcccccccc@{}}
\toprule
\multirow{2}{*}{Model} & \multicolumn{2}{c}{MNIST} & \multicolumn{2}{c}{USPS} & \multicolumn{2}{c}{20NEWS} & \multicolumn{2}{c}{RCV1} \\ \cmidrule(l){2-9} 
 & ACC & NMI & ACC & NMI & ACC & NMI & ACC & NMI \\ \midrule
KM & 53.5$\pm$0.3 & 49.8$\pm$0.5 & 67.3$\pm$0.1 & 61.4$\pm$0.1 & 23.2$\pm$1.5 & 21.6$\pm$1.8 & {\ul 50.8$\pm$2.9} & {\ul \textbf{31.3$\pm$5.4}} \\
AE-KM & {\ul 80.8$\pm$1.8} & 75.2$\pm$1.1 & {\ul 72.9$\pm$0.8} & {\ul 71.7$\pm$1.2} & \textbf{49.0$\pm$2.9} & 44.5$\pm$1.5 & {\ul \textbf{56.7$\pm$3.6}} & {\ul \textbf{31.5$\pm$4.3}} \\ \midrule
\multicolumn{9}{c}{Deep clustering approaches without pretraining} \\ \midrule
DCN$^\text{np}$ & 34.8$\pm$3.0 & 18.1$\pm$1.0 & 36.4$\pm$3.5 & 16.9$\pm$1.3 & 17.9$\pm$1.0 & 9.8$\pm$0.5 & 41.3$\pm$4.0 & 6.9$\pm$1.8 \\
DKM$^\text{a}$ & {\ul 82.3$\pm$3.2} & {\ul 78.0$\pm$1.9} & {\ul \textbf{75.5$\pm$6.8}} & {\ul 73.0$\pm$2.3} & {\ul 44.8$\pm$2.4} & {\ul 42.8$\pm$1.1} & {\ul \textbf{53.8$\pm$5.5}} & {\ul \textbf{28.0$\pm$5.8}} \\ \midrule
\multicolumn{9}{c}{Deep clustering approaches with pretraining} \\ \midrule
DCN$^\text{p}$ & {\ul 81.1$\pm$1.9} & 75.7$\pm$1.1 & 73.0$\pm$0.8 & {\ul 71.9$\pm$1.2} & \textbf{49.2$\pm$2.9} & 44.7$\pm$1.5 & {\ul \textbf{56.7$\pm$3.6}} & {\ul \textbf{31.6$\pm$4.3}} \\
DKM$^\text{p}$ & {\ul \textbf{84.0$\pm$2.2}} & \textbf{79.6$\pm$0.9} & {\ul \textbf{75.7$\pm$1.3}} & \textbf{77.6$\pm$1.1} & \textbf{51.2$\pm$2.8} & \textbf{46.7$\pm$1.2} & {\ul \textbf{58.3$\pm$3.8}} & {\ul \textbf{33.1$\pm$4.9}} \\ \bottomrule
\end{tabular}
}
%\vskip -0.1in
\end{table*}

\begin{table*}[t!]
\centering
\caption{Clustering results of the DKM and IDEC methods. Performance is measured in terms of NMI and ACC (\%); higher is better. Each cell contains the average and standard deviation computed over 10 runs. Bold (resp. underlined) values correspond to results with no significant difference ($p > 0.05$) to the best approach with (resp. without) pretraining for each dataset/metric pair.}
\label{tab:deep-clust-results}
\vskip 0.15in
\scalebox{0.87}{
\begin{tabular}{@{}ccccccccc@{}}
\toprule
\multirow{2}{*}{Model} & \multicolumn{2}{c}{MNIST} & \multicolumn{2}{c}{USPS} & \multicolumn{2}{c}{20NEWS} & \multicolumn{2}{c}{RCV1} \\ \cmidrule(l){2-9} 
 & ACC & NMI & ACC & NMI & ACC & NMI & ACC & NMI \\ \midrule
\multicolumn{9}{c}{Deep clustering approaches without pretraining} \\ \midrule
IDEC$^\text{np}$ & 61.8$\pm$3.0 & 62.4$\pm$1.6 & 53.9$\pm$5.1 & 50.0$\pm$3.8 & 22.3$\pm$1.5 & 22.3$\pm$1.5 & {\ul \textbf{56.7$\pm$5.3}} & {\ul \textbf{31.4$\pm$2.8}} \\
DKM$^\text{a}$ & {\ul 82.3$\pm$3.2} & {\ul 78.0$\pm$1.9} & {\ul \textbf{75.5$\pm$6.8}} & {\ul 73.0$\pm$2.3} & {\ul 44.8$\pm$2.4} & {\ul 42.8$\pm$1.1} & {\ul 53.8$\pm$5.5} & {\ul 28.0$\pm$5.8} \\ \midrule
\multicolumn{9}{c}{Deep clustering approaches with pretraining} \\ \midrule
IDEC$^\text{p}$ & \textbf{85.7$\pm$2.4} & \textbf{86.4$\pm$1.0} & {\ul \textbf{75.2$\pm$0.5}} & 74.9$\pm$0.6 & 40.5$\pm$1.3 & 38.2$\pm$1.0 & {\ul \textbf{59.5$\pm$5.7}} & {\ul \textbf{34.7$\pm$5.0}} \\
DKM$^\text{p}$ & {\ul \textbf{84.0$\pm$2.2}} & 79.6$\pm$0.9 & {\ul \textbf{75.7$\pm$1.3}} & \textbf{77.6$\pm$1.1} & \textbf{51.2$\pm$2.8} & \textbf{46.7$\pm$1.2} & {\ul \textbf{58.3$\pm$3.8}} & {\ul \textbf{33.1$\pm$4.9}} \\ \bottomrule
\end{tabular}
}
%\vskip -0.1in
\end{table*}

%\hspace{-0.3cm}
\subsection{Clustering results}

The results for the evaluation of the $k$-Means-related clustering methods on the different benchmark datasets are summarized in Table~\ref{tab:km-results}. The clustering performance is evaluated with respect to two standard measures~\cite{Cai2011}: Normalized Mutual Information (NMI) and the clustering accuracy (ACC). We report for each dataset/method pair the average and standard deviation of these metrics computed over 10 runs and conduct significance testing as previously described in the experimental protocol. The bold (resp. underlined) values in each column of Table~\ref{tab:km-results} correspond to results with no statistically significant difference ($p > 0.05$) to the best result with (resp. without) pretraining for the corresponding dataset/metric.

We first observe that when no pretraining is used, DKM with annealing (DKM$^\text{a}$) markedly outperforms DCN$^\text{np}$ on all datasets. DKM$^\text{a}$ achieves clustering performance similar to that obtained by pretraining-based methods. This confirms our intuition that the proposed annealing strategy can be seen as an alternative to pretraining. 

Among the approaches integrating representation learning with pretraining, the AE-KM method, that separately performs dimension reduction and $k$-Means clustering, overall obtains the worst results. This observation is in line with prior studies \cite{Yang2017,Guo2017b} and underlines again the importance of jointly learning representations and clustering. We note as well that, apart from DKM$^\text{a}$, pretraining-based deep clustering approaches substantially outperform their non-pretrained counterparts, which stresses the importance of pretraining.

Furthermore, DKM$^\text{p}$ yields significant improvements on all collections except RCV1 over DCN$^\text{p}$, the other ``true'' deep $k$-Means approach. In all cases, DCN$^\text{p}$ shows performance on par with that of AE-KM. This places, to the best of our knowledge, DKM$^\text{p}$ as the current best deep $k$-Means clustering method.

To further confirm DKM's efficacy, we also compare it against IDEC, a state-of-the-art deep clustering algorihm which is not based on $k$-Means. We report the corresponding results in Table~\ref{tab:deep-clust-results}. Once again, DKM$^\text{a}$ significantly outperforms its non-pretrained counterpart, IDEC$^\text{np}$, except on RCV1. We note as well that, with the exception of the NMI results on MNIST, DKM$^\text{p}$ is always either significantly better than IDEC$^\text{p}$ or with no significant difference from this latter. This shows that the proposed DKM is not only the strongest $k$-Means-related clustering approach, but is also remarkably competitive wrt deep clustering state of the art.

\subsection{Illustration of learned representations} 
\begin{figure}[t!]
\centering
\subfigure[AE]{\includegraphics[scale=0.14]{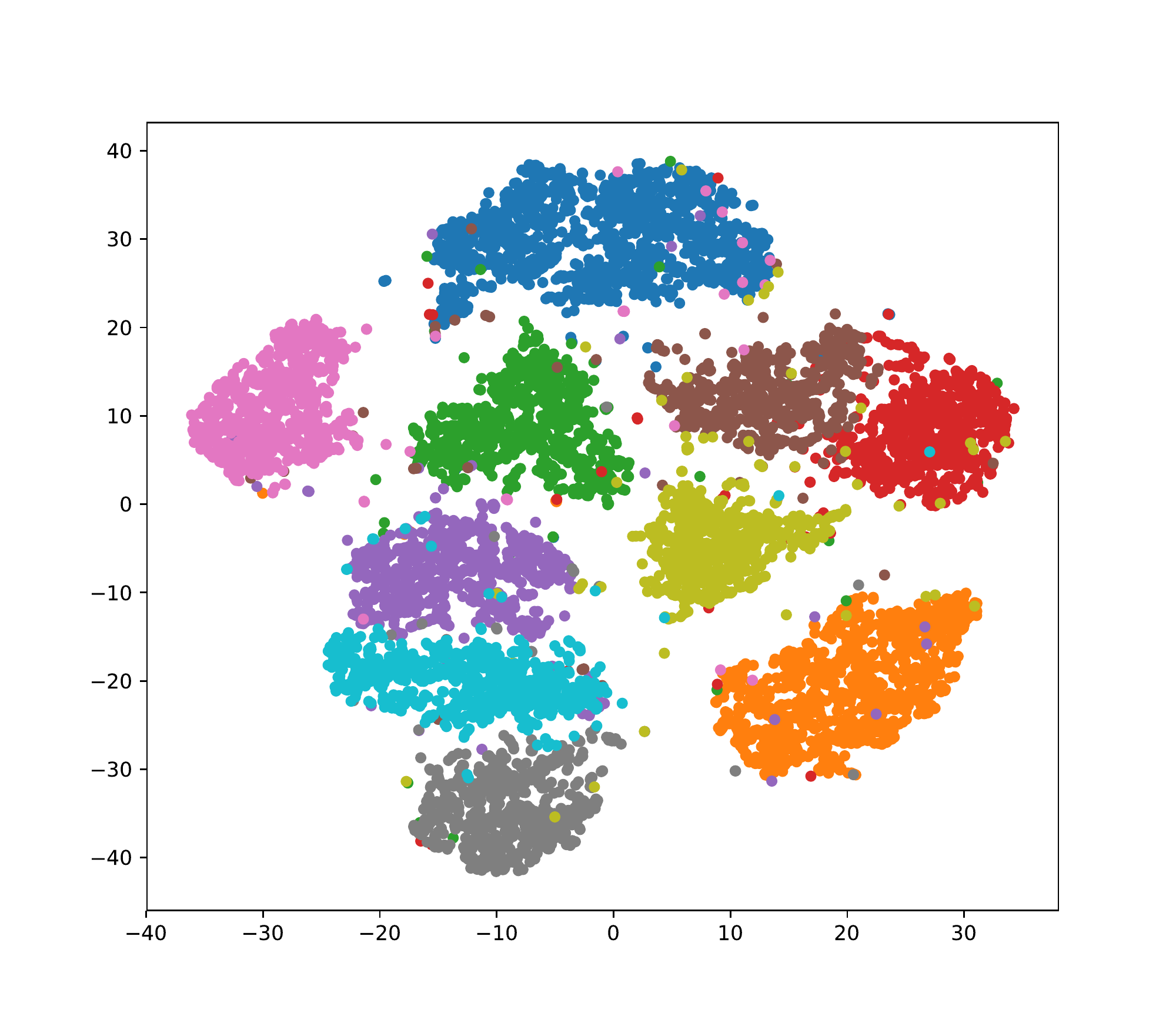}}
\quad
\subfigure[DCN$^\text{p}$]{\includegraphics[scale=0.14]{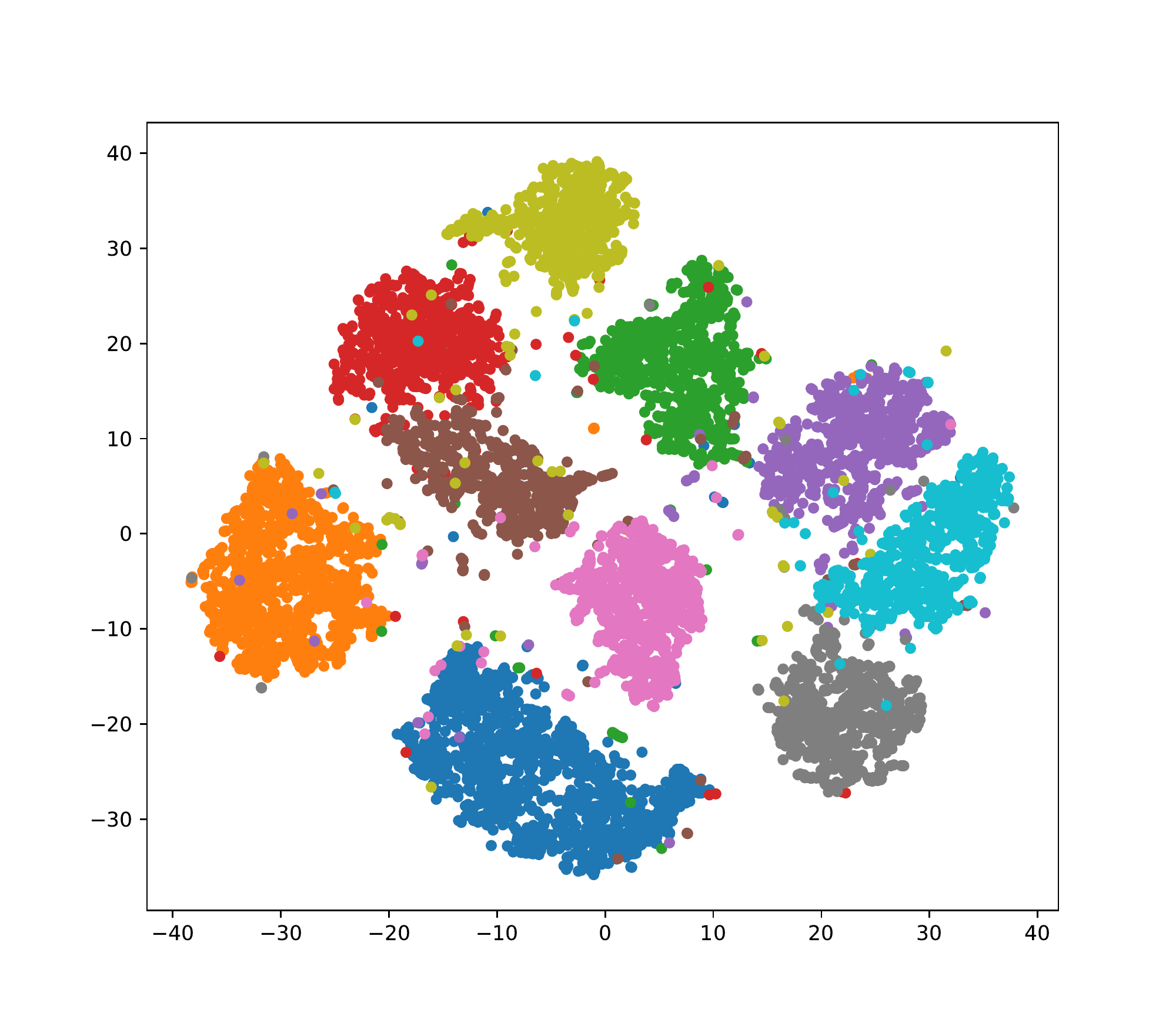}}
\quad
\subfigure[DKM$^\text{a}$]{\includegraphics[scale=0.14]{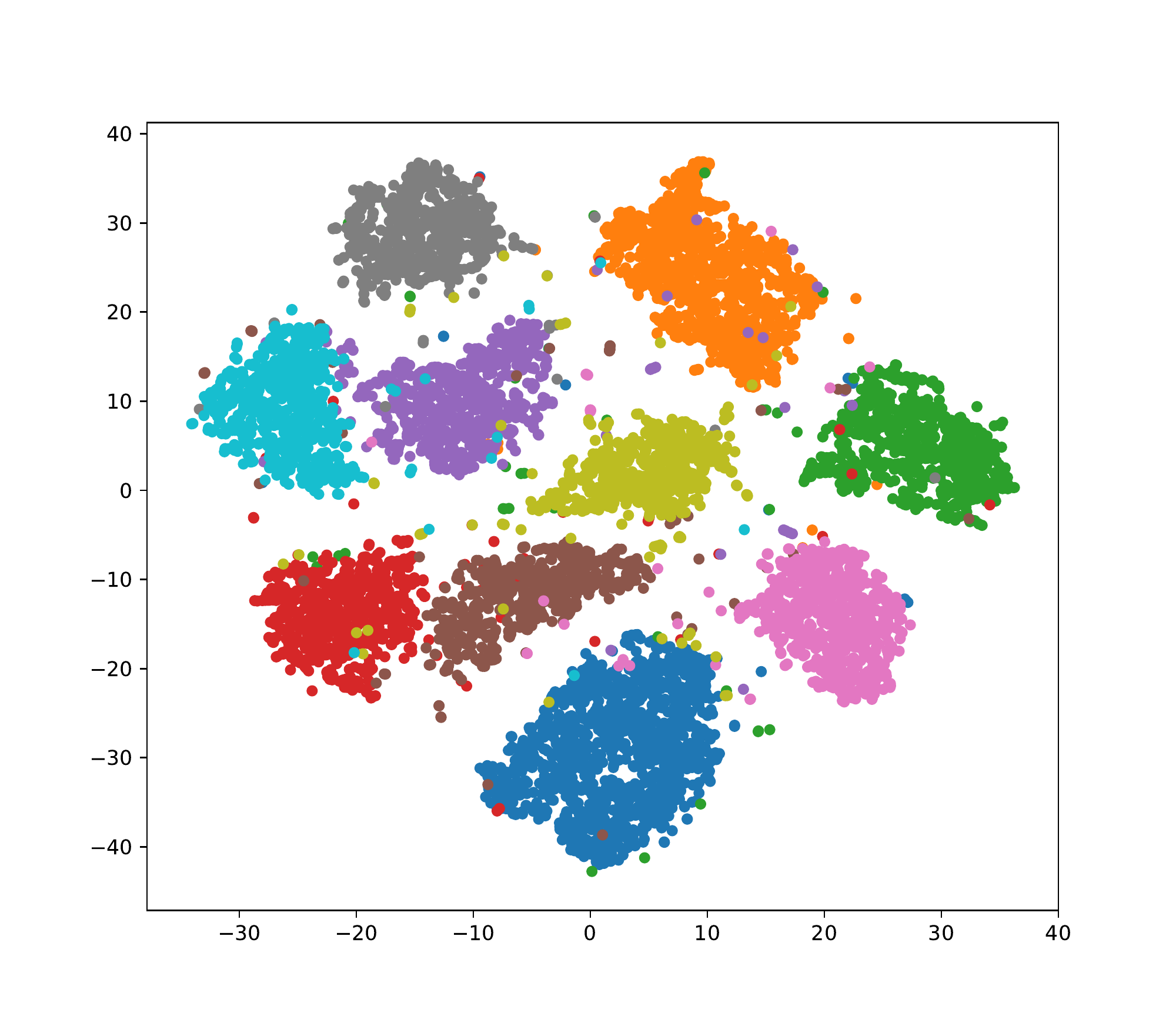}}
\quad
\subfigure[DKM$^\text{p}$]{\includegraphics[scale=0.14]{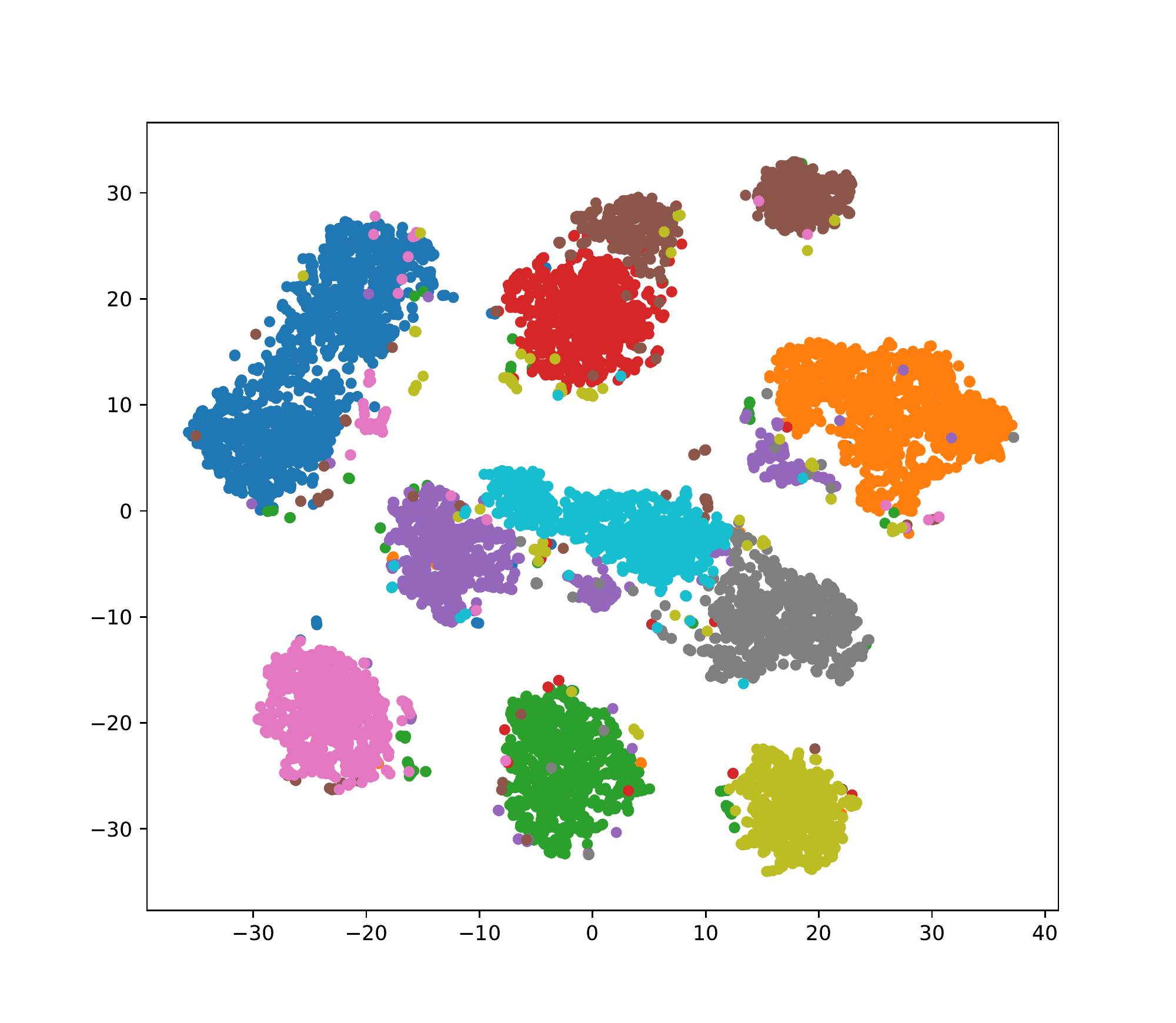}}
\caption{t-SNE visualization of the embedding spaces learned on USPS.}
\label{fig:t-sne}
\end{figure}

%In the previous experiment, we investigated the clustering ability of the different models. 
While the quality of the clustering results and that of the representations learned by the models are likely to be correlated, it is relevant to study to what extent learned representations are distorted to facilitate clustering.
To provide a more interpretable view of the representations learned by $k$-means-related deep clustering algorithm, we illustrate the embedded samples provided by AE (for comparison), DCN$^\text{p}$, DKM$^\text{a}$, and DKM$^\text{p}$ on USPS in Figure~\ref{fig:t-sne} (best viewed in color). DCN$^\text{np}$ was discarded due to its poor clustering performance. We used for that matter the $t$-SNE visualization method~\cite{vanderMaaten2008} to project the embeddings into a 2D space. We observe that the representations for points from different clusters are clearly better separated and disentangled in DKM$^\text{p}$ than in other models. This brings further support to our experimental results, which showed the superior ability of DKM$^\text{p}$ to learn representations that facilitate clustering.

\section{Conclusion}
\label{sec:concl}

We have presented in this paper a new approach for jointly clustering with $k$-Means and learning representations by considering the $k$-Means clustering loss as the limit of a differentiable function. To the best of our knowledge, this is the first approach that truly jointly optimizes, through simple stochastic gradient descent updates, representation and $k$-Means clustering losses. In addition to pretraining, that can be used in all methods, this approach can also rely on a deterministic annealing scheme for parameter initialization.
% ) can make use of an annealing scheme to dispenses with cumbersome pre-training procedures. Indeed, the annealing scheme used here reformulates the (generalized) $k$-Means objective function as the limit of a differentiable function and diminishes the influence of the initialization procedure.

We further conducted careful comparisons with previous approaches by ensuring that the same architecture, initialization and minibatches are used. The experiments conducted on several datasets confirm the good behavior of Deep $k$-Means that outperforms DCN, the current best approach for $k$-Means clustering in embedding spaces, on all the collections considered.

\section*{Acknowledgments}

Funding: This work was supported by the French National Agency for Research through the LOCUST project [grant number ANR-15-CE23-0027]; France's Auvergne-Rhône-Alpes region through the AISUA project [grant number 17011072 01 - 4102]. Declarations of interest: none.

\small

\bibliographystyle{abbrv}
\bibliography{deep-kmeans}

\appendix

\section{Proof of Property 1}

For $\displaystyle G_{k,f}(\vect{h}_{\theta}(x),\alpha;\mathcal{R}) = \frac{e^{-\alpha f(\vect{h}_{\theta}(x),\mathbf{r}_k)}}{\sum_{k'=1}^K e^{-\alpha f(\vect{h}_{\theta}(x),\mathbf{r}_{k'})}}$, we remind the following property:
\begin{property}(condition \textit{(ii)})\label{prop-f-alpha} If $c_f(\vect{h}_{\theta}(x);\mathcal{R})$ is unique for all $x \in \mathcal{X}$, then:
\begin{equation}\label{eq:propG1}
\underset{\alpha \rightarrow +\infty}{\lim} \, G_{k,f}(\vect{h}_{\theta}(x),\alpha;\mathcal{R}) = 
\begin{cases}
1 \, \ifn \, \mathbf{r}_k = c_f(\vect{h}_{\theta}(x);\mathcal{R}) \nonumber \\
0 \, \otherwise \nonumber
\end{cases}
\end{equation}
\end{property}
\textbf{Proof:} Let $\mathbf{r}_j = c_f(\vect{h}_{\theta}(x);\mathcal{R})$ and let us assume that $f$ is a distance. One has:
\begin{align}\label{eq:proof1}
\sum_{k'=1}^K & e^{- \alpha f(\vect{h}_{\theta}(x),\mathbf{r}_{k'})} = e^{- \alpha f(\vect{h}_{\theta}(x),\mathbf{r}_j)} \times \left( 1+ \sum_{k' \ne j} e^{-\alpha (f(\vect{h}_{\theta}(x),\mathbf{r}_{k'}) - f(\vect{h}_{\theta}(x),\mathbf{r}_j))} \right) \nonumber
\end{align}
%\begin{align}\label{eq:proof1}
%\sum_{k'=1}^K e^{- \alpha f(\vect{h}_{\theta}(x),\mathbf{r}_{k'})} = e^{- \alpha f(\vect{h}_{\theta}(x),\mathbf{r}_j)} \nonumber\times \left( 1+ \sum_{k' \ne j} e^{-\alpha (f(\vect{h}_{\theta}(x),\mathbf{r}_{k'}) - f(\vect{h}_{\theta}(x),\mathbf{r}_j))} \right) \nonumber
%\end{align}
%
As $f(\vect{h}_{\theta}(x),\mathbf{r}_j) < f(\vect{h}_{\theta}(x),\mathbf{r}_{k'}), \, \forall k' \ne j$, one has:
\[
\underset{\alpha \rightarrow +\infty}{\lim} e^{-\alpha (f(\vect{h}_{\theta}(x),\mathbf{r}_{k'}) - f(\vect{h}_{\theta}(x),\mathbf{r}_j))} =0
\]
Thus $\underset{\alpha \rightarrow +\infty}{\lim} G_{k,f}(\vect{h}_{\theta}(x),\alpha;\mathcal{R}) = 0$ if $k \ne j$ and $1$ if $k=j$. $\QEDB$

\section{Alternative choice for ${G_{k,f}}$}

As $G_{k,f}(\vect{h}_{\theta}(x),\alpha;\mathcal{R})$ plays the role of a closeness function for an object $x$ wrt representative $\vect{r}_k$, membership functions used in fuzzy clustering are potential candidates for $G_{k,f}$. In particular, the membership function of the fuzzy $C$-Means algorithm \cite{Bezdek1984} is a valid candidate according to conditions $(i)$ and $(ii)$. It takes the following form:
\[
G_{k,f}(\vect{h}_{\theta}(x),\alpha;\mathcal{R}) = \left( \sum_{k'=1}^K \left(\frac{f(\vect{h}_{\theta}(x),\mathbf{r}_k)}{f(\vect{h}_{\theta}(x),\mathbf{r}_{k'})}\right)^{\frac{2}{\alpha-1}} \right)^{-1}
\]
with $\alpha$ defined on $(1;+\infty)$ and $\alpha_0$ (condition $(ii)$) equal to 1. However, in addition to being slightly more complex than the parametrized softmax, this formulation presents the disadvantage that it may be undefined when a representative coincides with an object; another assumption (in addition to the uniqueness assumption) is required here to avoid such a case.

\section{Annealing scheme for $\alpha$ in DKM$^\text{a}$}

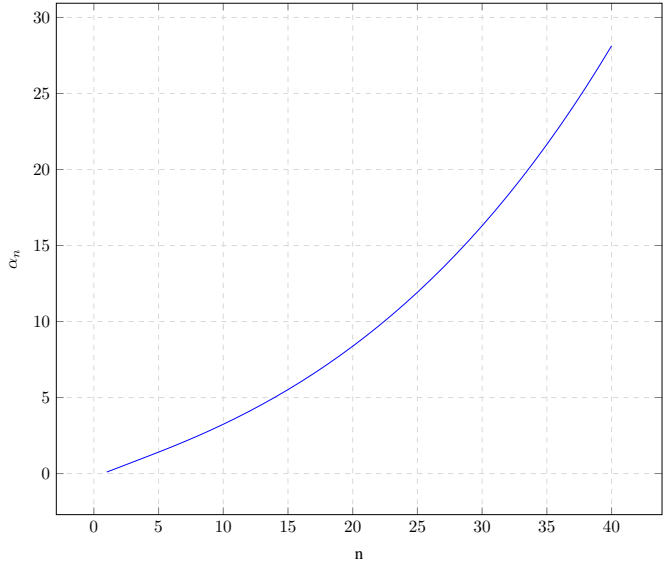
\begin{figure}
\centering
	\scalebox{0.65}{
    \begin{tikzpicture}
      \begin{axis}[
          width=1\linewidth, % Scale the plot to \linewidth
          grid=major, % Display a grid
          grid style={dashed,gray!30}, % Set the style
          xlabel=n, % Set the labels
          ylabel=$\alpha_n$,
          x tick label style={anchor=north}, % Display labels sideways
          y label style={at={(0.03,0.5)}},
          x label style={at={(0.5,-0.01)}},
        ]
        \addplot[color=blue]
        % add a plot from table; you select the columns by using the actual name in the .csv file (on top)
        table[x expr=\coordindex+1,y=alpha,col sep=comma] {data/dkm_annealed_alpha.csv}; 
      \end{axis}
    \end{tikzpicture}
    }
    %\vspace{-0.3cm}
\caption{Annealing scheme for inverse temperature $\alpha$, following the sequence $\alpha_{n+1} = 2^{1/\log(n)^2} \times \alpha_{n}$; $\alpha_1 = 0.1$.}
\label{fig:alpha_scheme}
\end{figure}

The scheme we used for the evolution of the inverse temperature $\alpha$ in DKM$^\text{a}$ is given by the following recursive sequence: $\alpha_{n+1} = 2^{1/\log(n)^2} \times \alpha_{n}$ with $m_{\alpha} = \alpha_1 = 0.1$. The 40 first terms of $(\alpha_{n})$ are plotted in Figure~\ref{fig:alpha_scheme}.

\section{Evaluation measures}

In our experiments, the clustering performance of the evaluated methods is evaluated with respect to two standard measures~\cite{Cai2011}: Normalized Mutual Information (NMI) and the clustering accuracy (ACC). NMI is an information-theoretic measure based on the mutual information of the ground-truth classes and the obtained clusters, normalized using the entropy of each.
Formally, let $S = \{S_1, \ldots, S_K\}$ and $C = \{C_1, \ldots, C_K\}$ denote the ground-truth classes and the obtained clusters, respectively. $S_i$ (resp. $C_j$) is the subset of data points from class $i$ (resp. cluster $j$). Let $N$ be the number of points in the dataset. The NMI is computed according to the following formula:
\begin{equation*}
	\text{NMI}(C, S) = \frac{I(C, S)}{\sqrt{H(C)\times H(S)}}
\end{equation*}
where $I(C, S) = \sum_{i, j} \frac{|C_i \cap S_j|}{N} \log \frac{N |C_i \cap S_j|}{|C_i| |S_j|}$ corresponds to the mutual information between the partitions $C$ and $S$, and $H(C) = - \sum_i \frac{|C_i|}{N} \log \frac{|C_i|}{N}$ is the entropy of $C$.
\vspace{0.3cm}

On the other hand, ACC measures the proportion of data points for which the obtained clusters can be correctly mapped to ground-truth classes, where the matching is based on the Hungarian algorithm~\cite{Kuhn1955}.
Let $s_i$ and $c_i$ further denote the ground-truth class and the obtained cluster, respectively, to which data point $x_i$, $i \in \{1, \ldots, N\}$ is assigned. Then the clustering accuracy is defined as follows:
\begin{equation*}
	\text{ACC}(C, S) = \max_{\phi} \frac{1}{N} \sum_{i=1}^N \mathbb{I}\{s_i = \phi(c_i)\}
\end{equation*}
where $\mathbb{I}$ denotes the indicator function: $\mathbb{I}\{\text{true}\} = 1$ and $\mathbb{I}\{\text{false}\} = 0$; $\phi$ is a mapping from cluster labels to class labels.
\vspace{0.3cm}

We additionally report in this supplementary material the clustering performance wrt to the adjusted Rand index (ARI) \cite{Vinh2010}. ARI counts the pairs of data points on which the classes and clusters agree or disagree, and is corrected for chance. Formally, ARI is given by:
\begin{equation*}
	\text{ARI}(C, S) = \frac {
		\sum_{ij}{\binom{|C_i \cap S_j|}{2}} - \binom{N}{2}^{-1}\sum_{i}{\binom{|C_i|}{2}}\sum_{j}{\binom{|S_j|}{2}}
	}{
		\frac{1}{2}\left(\sum_{i}{\binom {|C_i|}{2}}+\sum _{j}{\binom{|S_j|}{2}}\right) - \binom{N}{2}^{-1} \sum_{i}{\binom{|C_i|}{2}}\sum _{j}{\binom{|S_j|}{2}} 
	}
\end{equation*}
%}

%\section{Deep \textit{$k$}-means Pretrained Annealed ( DKM$^\text{pa}$)}
%In paper we showed the possibility of considering different versions of DKM which are  DKM$^\text{a}$ and  DKM$^\text{p}$. Another possibility is to combine the characteristics of  DKM$^\text{a}$ and   DKM$^\text{p}$. Indeed, we can use pretraining and annealing at the same time. This version of DKM is called  DKM$^\text{pa}$ which due to lack of space in the paper, we give its results later in this supplementary material. 

\section{Dataset statistics and optimal hyperparameters}

We summarize in Table~\ref{tab:settings} the statistics of the different datasets used in the experiments, as well as the dataset-specific optimal values of the hyperparameter ($\lambda$ for DKM-based and DCN-based methods and $\gamma$ for IDEC-based ones) which trades off between the reconstruction loss and the clustering loss. We remind that this optimal value was determined using a validation set, disjoint from the test set on which we reported results in the paper.

\begin{table}[t]
\centering
\caption{Statistics of the datasets and dataset-specific optimal trade-off hyperparameters ($\lambda$ for DKM-based and DCN-based methods and $\gamma$ for IDEC-based ones) determined on the validation set.}
\label{tab:settings}
\vskip 0.15in
\scalebox{1}{
\begin{tabular}{@{}lcccc@{}}
\toprule
Dataset & MNIST & USPS & 20NEWS & RCV1 \\ \midrule
\#Samples & 70,000 & 9,298 & 18,846 & 10,000 \\
\#Classes & 10 & 10 & 20 & 4 \\
Dimensions & 28 $\times$ 28 & 16 $\times$ 16 & 2,000 & 2,000 \\ \midrule
$\lambda_{\text{DKM}^\text{a}}$ & 1e-1 & 1e-1 & 1e-4 & 1e-4 \\
$\lambda_{\text{DKM}^\text{p}}$ & 1e+0 & 1e+0 & 1e-1 & 1e-2 \\
$\lambda_{\text{DCN}^\text{p}}$ & 1e+1 & 1e-1 & 1e-1 & 1e-1 \\
$\lambda_{\text{DCN}^\text{np}}$ & 1e-2 & 1e-1 & 1e-4 & 1e-3 \\
$\gamma_{\text{IDEC}^\text{p}}$ & 1e-2 & 1e-3 & 1e-1 & 1e-3 \\
$\gamma_{\text{IDEC}^\text{np}}$ & 1e-3 & 1e-1 & 1e-3 & 1e-4 \\ \bottomrule
\end{tabular}
}
\vskip -0.1in
\end{table}

\section{Additional results}

The additional results given in this section have also been computed from 10 seeded runs whenever possible and Student's $t$-test was performed from those 10 samples.

\subsection{ARI results}

We report in Table~\ref{tab:ari-km-results} the results obtained by $k$-Means-related methods wrt the ARI measure on the datasets used in the paper. Similarly, Table~\ref{tab:ari-deep-clust-results} compares the results of the approaches based on DKM and IDEC in terms of ARI.

\begin{table}[t]
\centering
\caption{Clustering results of the $k$-Means-related methods. Performance is measured in terms of ARI (\%); higher is better. Each cell contains the average and standard deviation computed over 10 runs. Bold (resp. underlined) values correspond to results with no significant difference ($p > 0.05$) to the best approach with (resp. without) pretraining for each dataset/metric pair.}
\label{tab:ari-km-results}
\vskip 0.15in
\scalebox{1}{
\begin{tabular}{@{}ccccc@{}}
\toprule
Model & MNIST & USPS & 20NEWS & RCV1 \\ \midrule
KM & 36.6$\pm$0.1 & 53.5$\pm$0.1 & 7.6$\pm$0.9 & {\ul 20.6$\pm$2.8} \\
AE-KM & 69.4$\pm$1.8 & {\ul 63.2$\pm$1.5} & 31.0$\pm$1.6 & \textbf{23.9$\pm$4.3} \\ \midrule
\multicolumn{5}{c}{Deep clustering approaches without pretraining} \\ \midrule
DCN$^\text{np}$ & 15.6$\pm$1.1 & 14.7$\pm$1.8 & 5.7$\pm$0.5 & 6.9$\pm$2.1 \\
DKM$^\text{a}$ & {\ul \textbf{73.6$\pm$3.1}} & {\ul \textbf{66.3$\pm$4.9}} & {\ul 26.7$\pm$1.5} & {\ul 20.7$\pm$4.4} \\ \midrule
\multicolumn{5}{c}{Deep clustering approaches with pretraining} \\ \midrule
DCN$^\text{p}$ & 70.2$\pm$1.8 & {\ul 63.4$\pm$1.5} & 31.3$\pm$1.6 & \textbf{24.0$\pm$4.3} \\
DKM$^\text{p}$ & {\ul \textbf{75.0$\pm$1.8}} & {\ul \textbf{68.5$\pm$1.8}} & \textbf{33.9$\pm$1.5} & \textbf{26.5$\pm$4.9} \\ \bottomrule
\end{tabular}
}
%\vskip -0.1in
\end{table}

\begin{table}[t]
\centering
\caption{Clustering results of the DKM and IDEC methods. Performance is measured in terms of ARI (\%); higher is better. Each cell contains the average and standard deviation computed over 10 runs. Bold (resp. underlined) values correspond to results with no significant difference ($p > 0.05$) to the best approach with (resp. without) pretraining for each dataset/metric pair.}
\label{tab:ari-deep-clust-results}
\vskip 0.15in
\scalebox{1}{
\begin{tabular}{@{}ccccc@{}}
\toprule
Model & MNIST & USPS & 20NEWS & RCV1 \\ \midrule
\multicolumn{5}{c}{Deep clustering approaches without pretraining} \\ \midrule
IDEC$^\text{np}$ & 49.1$\pm$3.0 & 40.2$\pm$5.1 & 9.8$\pm$1.5 & {\ul \textbf{28.5$\pm$5.3}} \\
DKM$^\text{a}$ & {\ul 73.6$\pm$3.1} & {\ul \textbf{66.3$\pm$4.9}} & {\ul 26.7$\pm$1.5} & 20.7$\pm$4.4 \\ \midrule
\multicolumn{5}{c}{Deep clustering approaches with pretraining} \\ \midrule
IDEC$^\text{p}$ & \textbf{81.5$\pm$2.4} & {\ul \textbf{68.1$\pm$0.5}} & {\ul 26.0$\pm$1.3} & {\ul \textbf{32.9$\pm$5.7}} \\
DKM$^\text{p}$ & {\ul 75.0$\pm$1.8} & {\ul \textbf{68.5$\pm$1.8}} & \textbf{33.9$\pm$1.5} & {\ul 26.5$\pm$4.9} \\ \bottomrule
\end{tabular}
}
%\vskip -0.1in
\end{table}

\subsection{Cosine distance}

The proposed Deep $k$-Means framework enables the use of different distance and similarity functions to compute the AE's reconstruction error (based on function $g$) and the clustering loss (based on function $f$). In the paper, we adopted the euclidean distance for both $f$ and $g$. For the sake of comprehensiveness, we performed additional experiments on DKM using different such functions. In particular, we showcase in Table~\ref{tab:cosine_results} the results obtained by choosing the cosine distance for $f$ and the euclidean distance for $g$ (\textbf{DKM$^\text{a}_\text{ce}$} and \textbf{DKM$^\text{p}_\text{ce}$}) in comparison to using euclidean distance for both $f$ and $g$ (\textbf{DKM$^\text{a}_\text{ee}$} and \textbf{DKM$^\text{p}_\text{ee}$})~-- these latter corresponding to the approaches reported in the paper. Note that for DKM$^\text{a}_\text{ce}$ and DKM$^\text{p}_\text{ce}$ as well the reported results correspond to those obtained with the optimal lambda determined on the validation set.

\begin{table}[t]
\centering
\caption{Clustering results for DKM using the euclidean and cosine distances on 20NEWS. Each cell contains the average and standard deviation computed over 10 runs. Bold values correspond to results with no significant difference ($p > 0.05$) to the best for each dataset/metric pair..}
\label{tab:cosine_results}
\vskip 0.15in
\scalebox{1}{
\begin{tabular}{@{}cccc@{}}
\toprule
Model & ACC & NMI & ARI \\ \midrule
\multicolumn{4}{c}{$f$ = euclidean distance, $g$ = euclidean distance} \\ \midrule
DKM$^\text{a}_\text{ee}$ & 44.8$\pm$2.4 & 42.8$\pm$1.1 & 26.7$\pm$1.5 \\
DKM$^\text{p}_\text{ee}$ & \textbf{51.2$\pm$2.8} & {\textbf{46.7$\pm$1.2}} & {\textbf{33.9$\pm$1.5}} \\ \midrule
\multicolumn{4}{c}{$f$ = cosine distance, $g$ = euclidean distance} \\ \midrule
DKM$^\text{a}_\text{ce}$ & {\textbf{51.3$\pm$1.5}} & 44.4$\pm$0.7 & 32.6$\pm$1.0 \\
DKM$^\text{p}_\text{ce}$ & \textbf{51.0$\pm$2.6} & 45.1$\pm$1.2 & \textbf{33.0$\pm$1.1} \\ \bottomrule
\end{tabular}
}
\vskip -0.1in
\end{table}

\begin{table*}[t!]
\centering
\caption{Clustering results for $k$-Means applied to different learned embedding spaces to measure the $k$-Means-friendliness of each method. Performance is measured in terms of NMI and clustering accuracy (\%), averaged over 10 runs with standard deviation. Bold values correspond to results with no significant difference ($p > 0.05$) to the best for each dataset/metric.}
\label{tab:friendliness}
\vskip 0.15in
\scalebox{0.87}{
\begin{tabular}{@{}ccccccccc@{}}
\toprule
\multirow{2}{*}{\begin{tabular}[c]{@{}c@{}}Model\end{tabular}} & \multicolumn{2}{c}{MNIST} & \multicolumn{2}{c}{USPS} & \multicolumn{2}{c}{20NEWS} & \multicolumn{2}{c}{RCV1} \\ \cmidrule(l){2-9} 
 & ACC & NMI & ACC & NMI & ACC & NMI & ACC & NMI \\ \midrule
AE-KM & 80.8$\pm$1.8 & 75.2$\pm$1.1 & 72.9$\pm$0.8 & 71.7$\pm$1.2 & 49.0$\pm$2.9 & 44.5$\pm$1.5 & \textbf{56.7$\pm$3.6} & \textbf{31.5$\pm$4.3} \\
DCN$^\text{p}$ + KM & \textbf{84.9$\pm$3.1} & \textbf{79.4$\pm$1.5} & \textbf{73.9$\pm$0.7} & 74.1$\pm$1.1 & \textbf{50.5$\pm$3.1} & \textbf{46.5$\pm$1.6} & \textbf{57.3$\pm$3.6} & \textbf{32.3$\pm$4.4} \\
DKM$^\text{a}$ + KM & \textbf{84.8$\pm$1.3} & \textbf{78.7$\pm$0.8} & {\textbf{76.9$\pm$4.9}} & 74.3$\pm$1.5 & 49.0$\pm$2.5 & 44.0$\pm$1.0 & \textbf{53.4$\pm$5.9} & 27.4$\pm$5.3 \\
DKM$^\text{p}$ + KM & {\textbf{85.1$\pm$3.0}} & {\textbf{79.9$\pm$1.5}} & \textbf{75.7$\pm$1.3} & {\textbf{77.6$\pm$1.1}} & {\textbf{52.1$\pm$2.7}} & {\textbf{47.1$\pm$1.3}} & {\textbf{58.3$\pm$3.8}} & {\textbf{33.0$\pm$4.9}} \\ \bottomrule
\end{tabular}
}
%\vskip -0.1in
\end{table*}

\subsection{$k$-Means-friendliness of learned representations} 

In addition to the previous experiments~-- which evaluate the clustering ability of the different approaches~-- we analyzed how effective applying $k$-Means to the representations learned by DCN$^\text{p}$, DKM$^\text{a}$, and DKM$^\text{p}$ is in comparison to applying $k$-Means to the AE-based representations (\textit{i.e.}, AE-KM). In other words, we evaluate the ``$k$-Means-friendliness'' of the learned representations. The results of this experiment are reported in Table~\ref{tab:friendliness}. We can observe that on most datasets the representations learned by $k$-Means-related deep clustering approaches lead to significant improvement wrt AE-learned representations. This confirms that all these deep clustering methods truly bias their representations. 
%Interestingly, we note that applying $k$-Means to the representations learned by DCN$^\text{p}$ yields substantial improvements in comparison to the results reported in Table~\ref{tab:km-results}. 
Overall, although the difference is not statistically significant on all datasets/metrics, the representations learned by DKM$^\text{p}$ are shown to be the most appropriate to $k$-Means. This goes in line with the insight gathered from the previous experiments.

%\section{Appendix}

\end{document}